\documentclass[letterpaper, 10 pt, conference]{ieeeconf}  

\newcommand{\figref}[1]{\autoref{#1}}
\newcommand{\secref}[1]{Section~\ref{#1}}
\newcommand{\tabref}[1]{\autoref{#1}}
\newcommand{\heading}[1]{\smallskip\noindent\textbf{#1}}

\newcommand{\ie}{i.e.}
\newcommand{\eg}{e.g.}

\usepackage[hidelinks]{hyperref}
\usepackage{graphicx}
\usepackage{svg}
\usepackage{amsmath}
\usepackage{amsmath}
\usepackage{mathtools}
\usepackage{cite}

\usepackage{url} 
\usepackage[utf8]{inputenc}
\usepackage{array}

\usepackage{xcolor}
\usepackage{listings}
\usepackage{booktabs}
\usepackage{algorithm,algpseudocode}
\usepackage[labelformat=simple]{subcaption}
\usepackage{stfloats}

\usepackage{cleveref}

\usepackage{booktabs}
\usepackage[normalem]{ulem}

\definecolor{dkgreen}{rgb}{0,0.6,0}
\definecolor{gray}{rgb}{0.5,0.5,0.5}
\definecolor{mauve}{rgb}{0.58,0,0.82}

\usepackage{booktabs,multirow}
\usepackage{siunitx}
\usepackage{colortbl}%
  \newcommand{\myrowcolour}{\rowcolor[gray]{0.925}}

\usepackage{cancel}
\usepackage{soul}
\usepackage{xcolor}
\makeatletter
\newcommand{\stcolor}[2]{\def\SOUL@stcolor{\color{#1}}\st{#2}}
\makeatother
\definecolor{myrevise}{HTML}{048243}

\newcommand{\name}{{OmniVLA}} 

\title{\LARGE \bf 
\name{}: Physically-Grounded Multimodal VLA with Unified\\ Multi-Sensor Perception for Robotic Manipulation}

\begin{document}

\newcommand{\todoEdit}[1]{\textcolor{black}{#1}} 
\newcommand{\minorEdit}[1]{\textcolor{black}{#1}} 
\newcommand{\majorEdit}[1]{\textcolor{black}{#1}}  
\newcommand{\noteEdit}[1]{\textcolor{red}{#1}}  
\newcommand{\revisionEdit}[1]{\textcolor{black}{#1}}  

\newcommand{\affmark}[1]{\textsuperscript{#1}}
\newcommand{\coprime}{\textsuperscript{*}}
\newcommand{\corr}{\textsuperscript{\dag}}
\newcommand{\namecomma}{\unskip,\,~} 

\author{%
Heyu Guo\affmark{1}\coprime\namecomma
Shanmu Wang\affmark{2}\coprime\namecomma
Ruichun Ma\affmark{3}\corr\namecomma
Shiqi Jiang\affmark{3}\\
Yasaman Ghasempour\affmark{1}\namecomma
Omid Abari\affmark{2}\namecomma
Baining Guo\affmark{3}\namecomma
Lili Qiu\affmark{3}\\[6pt]
\affmark{1}Princeton University \enspace
\affmark{2}University of California, Los Angeles \enspace
\affmark{3}Microsoft Research Asia
}

\maketitle

\begingroup
\renewcommand\thefootnote{\fnsymbol{footnote}}
\footnotetext[1]{Co-primary authors; work completed during internships at Microsoft Research Asia.} 
\footnotetext[2]{Corresponding author (ruichunma@microsoft.com).} 
\endgroup


\begin{abstract}
Vision-language-action (VLA) models have shown strong generalization in robotic manipulation through large-scale vision-language pretraining. However, most existing models rely solely on RGB cameras, limiting their perception and, consequently, manipulation capabilities.
We present \textbf{\name{}}, an omni-modality VLA model that integrates novel sensing modalities to enable beyond-RGB robotic perception and manipulation.
The core of our approach is the \emph{sensor-masked image}, a unified representation that overlays physically meaningful, spatially grounded masks onto the RGB images.
These masks are derived from sensors including an infrared camera, a mmWave radar, and a microphone array. 
This image-native unification keeps sensor input close to RGB statistics to facilitate training, provides a uniform interface across sensor hardware, and enables data-efficient learning with lightweight per-sensor projectors. 
\majorEdit{Building on this, we design a multimodal vision-language-action model architecture and train \name{} by extending an RGB-pretrained VLA backbone.}
We evaluate \name{} on challenging real-world tasks that require sensor-modality perception to guide the manipulation. 
\name{} achieves an average task success rate of 84\%, significantly outperforms both RGB-only and raw-sensor-input baseline models by {59\%} and {28\%} respectively, meanwhile showing higher learning efficiency and stronger generalization capability.
\end{abstract}



\section{Introduction}
\label{sec:intro}

Vision--language--action (VLA) models~\cite{black2024pi0,shukor2025smolvla} recently emerged as a powerful paradigm for generalist robotic policies. They leverage vision-language pretraining to map user prompts and camera observations to robot actions, showing great generalization and instruction following capability.
However, most VLA models are limited to RGB camera image input, which constrains their perception capabilities and prevents them from handling tasks that require non-RGB cues.
This undermines the potential of robots to utilize additional sensory hardware and perform challenging tasks that require perception capability similar to or even beyond humans.
For example,
Infrared (IR) cameras reveal temperature contrast for tasks such as search and rescue; millimeter wave (mmWave) radars penetrate occlusions, such as cardboard or clothing, with radio-frequency (RF) signals to localize hidden objects\majorEdit{~\cite{guan2020through, dodds2025non}}; acoustic microphone arrays enable human-like environmental awareness to ambient sound and react accordingly.
Adding per-sensor encoders or directly feeding raw sensor data to VLA models, however, increases system complexity, exacerbates data scarcity, and reduces compatibility with pretrained vision-language models.

\begin{figure}
    \centering
    \setlength{\belowcaptionskip}{-10pt}
    \includegraphics[width=0.98\columnwidth]{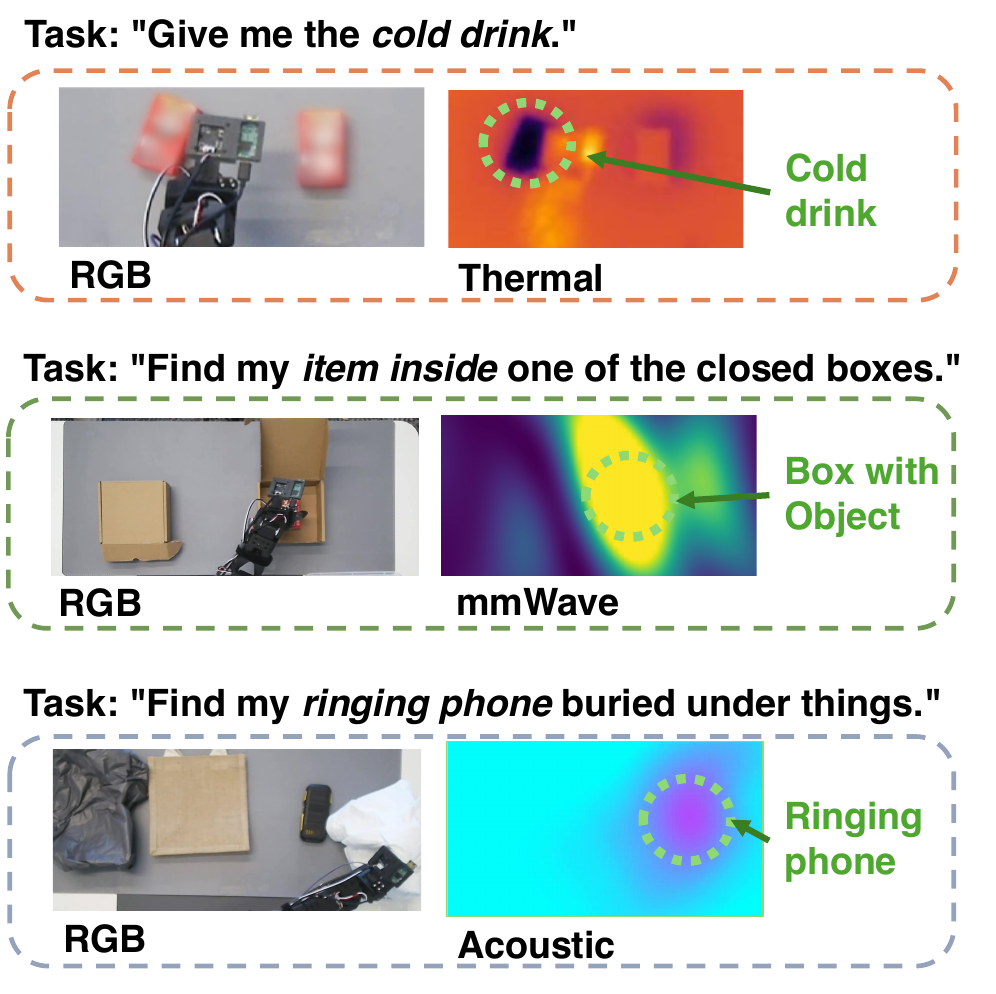}
    \caption{Instead of relying solely on RGB cameras, \name{} equips robots with multi-sensor perception. We use beamforming heatmaps as acoustic and mmWave sensor images to highlight sound source and hidden item respectively.}
    \label{fig:intro-sensor-motiv}
\end{figure}


\majorEdit{In this paper, we aim to equip VLA models with unified
multi-sensor perception, utilizing sensor modalities including infrared, acoustic, and mmWave. 
This enables robots to combine the strong generalization of foundation models and physical information from various sensors seamlessly, to enable physically-grounded spatial intelligence.}

There are several challenges in integrating diverse sensors with a VLA model. 
First, VLA models need to effectively interpret heterogeneous sensor information and use them to guide the action; naively feeding raw sensor streams leads to poor performance and data efficiency as shown in \secref{sec:eval}, because existing VLA backbones are trained primarily on RGB images.
Second, sensors differ in format, field of view, and resolution, calling for a scalable, uniform representation rather than training sensor fusion models that depend on specific hardware.
Lastly, sensor modality data are much more scarce compared to web-scale image-text data pairs, so we require a data-efficient approach for training.
To solve these challenges,
we take inspiration from how the human brain interprets sensor information: 
as we are used to RGB images, we naturally anchor other sensor cues to the RGB view—for example, interpreting infrared camera images by associating temperatures with objects visible in the image.

We present \textbf{\name{}}, the first multisensory VLA that integrates novel sensing modalities to enable beyond-RGB
robotic perception and manipulation by unifying heterogeneous sensors into an image-native space.
The core of our design is an intermediate representation, \emph{sensor-masked images}, which is produced by semantically segmenting the RGB image and overlaying the relevant sensor information as colored masks.
Such representation makes sure sensor information is spatially grounded and semantically aligned with the RGB image to ease the integration with VLA models.
This brings several benefits that solve the challenges above:
(i) making sensor information spatially grounded in RGB pixel coordinates to facilitate robotic manipulation on target objects, (ii) remaining close to RGB statistics so existing vision encoders and VLA backbones can be reused for further training, (iii) providing a uniform representation across sensors, resolutions, and hardware variants.


Building on the sensor-masked images, we propose a tailored VLA model architecture (\figref{fig:design-system-overview}).
We first convert all raw sensor measurements to image-like 2D spatial representations, specifically, we perform beamforming for mmWave radar and acoustic array data to acquire heatmap-like raw sensor images.
Then, to generate masks for interested objects in the scene, we use a cloud-based Vision-Language Model (VLM) to interpret task request and generate a prompt for grounded SAM~\cite{ren2024grounded} to provide semantic-based segmentation masks.
We further overlay sensor information on the RGB images at masked regions to acquire the sensor-masked images as the input to a frozen vision encoder.
After the vision encoder, we add lightweight projection layers for each sensor to generate better aligned tokens for sensor images.
\minorEdit{Finally, a LLM backbone processes the input tokens and conditions the diffusion-based action expert for the robot action output.}

We build a multi-sensor robot arm prototype to collect RGB camera and sensor data, paired with action demonstrations, and train \name{} with the collected datasets.
We evaluate \name{} extensively with several manipulation tasks that require sensor-modality guidance, including thermal-based pick-and-place, mmWave-based see-through boxes and opening the non-empty one, and acoustic-based uncovering ringing phone beneath clothes. 
\name{} achieves an average task success rate of 84\%, significantly outperforms 25\% success rate of RGB-only baseline and 56\% success rate of raw-sensor-input baseline. This shows the benefit from our unique sensing capabilities, and the performance gain from our sensor-masked image representation.
We also highlight the data efficiency of our approach by achieving similar success rate with only 50\% training data compared to raw-sensor-input baseline.
Moreover, we show that our approach provides strong generalization capability across three unseen tasks, outperforming base VLA model and raw-sensor-data based model by 59\% and 28\% respectively on average success rate. 


To summarize, we make the following contributions:
\begin{enumerate}
    \item 
    To our knowledge, \name{} is the first VLA model that unifies multiple sensing modalities, including infrared, mmWave, acoustic, to enable robotic manipulation tasks beyond RGB-based perception capability.
    \item 
    We introduce \emph{sensor-masked images}, a spatially grounded and semantically aligned representation that allows reusing pre-trained vision encoders, provides a uniform representation across sensor hardware, and improves learning efficiency.
    \item 
    We present a lightweight \name{} model architecture and evaluate system performance with extensive experimental evaluation. 
    \todoEdit{We open-source the project at \href{https://github.com/GuoHeyu/OmniVLA}{https://github.com/GuoHeyu/OmniVLA}.}
\end{enumerate}
\section{Related Work}
\label{sec:related}

\subsection{Vision-Language-Action Models}
Vision-language-action (VLA) models have been a popular research paradigm for robotic manipulation, using the language prompt and video feed as input and generating robot actions in an end-to-end manner. \minorEdit{Conventional robotic manipulation policies using reinforcement learning and simulators provide great performance on specific tasks \cite{chebotar2023q, luo2023action, ying2024peac, kumar2021workflow, chebotar2021actionable,kumar2022pre}, while VLA models have shown great improvement in few-shot task generalization and instruction following by leveraging web-scale pretraining.} The majority of VLA models only take video from RGB cameras as visual input \cite{mees2024octo, kim2024openvla, wen2025dexvla, ze2024generalizable, black2024pi0,shukor2025smolvla}. However, these works are inherently limited by the RGB camera input and unable to finish tasks that require perception capabilities beyond RGB.

To address this issue, researchers propose novel Vision-Language-Action models with additional sensor input. Depth information is widely employed to enhance the capability of VLA models with better spatial-temporal understanding \cite{li2025pointvla, bhat20253d, qu2025spatialvla, patratskiy2025spatial, zhen20243d}. Other works incorporate tactile perception in VLA models for better task planning, grounding and reasoning capabilities \cite{bi2025vla, huang2025tactile}. Moreover, Vlas~\cite{zhao2025vlas} integrates speech information for convenient and personalized human-robot interaction. 
\minorEdit{MultiPLY~\cite{hong2024multiply} introduces an embodied LLM for planning multisensory interactions with the environment, but it is based on simulator only and cannot generate actions for manipulation tasks.}

Prior works have not explored the integration of novel sensing modalities like thermal, acoustic, and mmWave, which provide unique sensing information for manipulation tasks.
Also, existing approaches often require training sensor encoders for each modality, which needs a large amount of data. In contrast, we propose a data-efficient approach towards beyond-RGB perception. 
Moreover, previous works propose complex and specific model architecture for extra sensor and image fusion, which is not generalizable to diverse sensors. 
In contrast, we propose sensor-masked images, a simple and unified sensor fusion strategy for VLA models.

\subsection{Multi-sensor Fusion based Perception}

Multi-sensor fusion based perception has received significant interest recently, especially for 3D detection and other downstream tasks for autonomous driving. The benefits for multi-sensor fusion are comprehensive environmental understanding for complementary sensors \cite{liu2022bevfusion}. One typical type of multi-sensor fusion is mmWave radar and camera sensor fusion, widely used in autonomous driving because of its reliability in rainy and foggy environments \cite{guan2020through, wolters2024unleashing, Lin_2024_CVPR, palladin2024samfusion, kim2024crt, xiong2025lxlv2}. 
Another thread of recent work fuses RF signals for robotic perception to achieve non-line-of-sight object finding~\cite{boroushaki2021rfusion, boroushaki2021robotic, boroushaki2022fusebot}, but it requires extra RFID tags attached on the objects and are designed only for RF signals. 


Prior works focus on environmental perception tasks like 3D object detection, semantic scene understanding tasks~\cite{Wang_2024_CVPR, han2025multimodal}, while we aim for robot manipulation by generating actions.
These prior works also heavily rely on model architectures tailored for domain-specific downstream tasks with specific sensors, unable to provide strong generalization or instruction following of VLA models.
Moreover, \name{} provides a general framework for more diverse sensors. 


\section{System Design}
\label{sec:design}

\begin{figure*}[t]
    \includegraphics[width=0.99\linewidth]{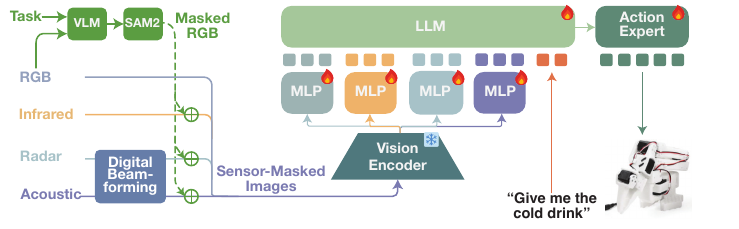}
    \vspace{-10pt}
    \caption{System Overview. \name{} processes diverse sensor data into image-like 2D spatial representations, and then overlays sensor information onto RGB images to produce spatially grounded and semantically aligned \textit{sensor-masked images}. We train \name{} by extending an RGB-pretrained VLA backbone to perform challenging tasks beyond RGB perception.}
    \label{fig:design-system-overview}
\end{figure*}

\subsection{System Overview}
\name{} contains two parts: sensor-masked image generation and multi-sensor vision-language-action model architecture. The first part first extracts raw sensor data, then preprocesses it into 2D sensor images/heatmaps. After that, we use a semantic segmentation model to generate the masks from the RGB images following a VLM-generated prompt. Then we overlay the sensor images on the masked regions of RGB images to output sensor-masked images, which are the input for our multi-sensor vision-language-action model. 
The second part is a multi-sensor vision-language-action model backbone that is designed for sensor-masked images to capture each sensor's input and avoid requiring a large amount of sensor data for training. 
We utilize the existing frozen vision encoders to encode sensor-masked images. 
For each sensor modality, we use individual multi-layer perceptron layers to align sensor image tokens with language and RGB image tokens. 
The tokens are concatenated together with language tokens as input for the large language model in the architecture, and then we generate the final action predictions using the action expert.

\begin{figure*}[t]
    \centering
    \includegraphics[width=0.99\linewidth]{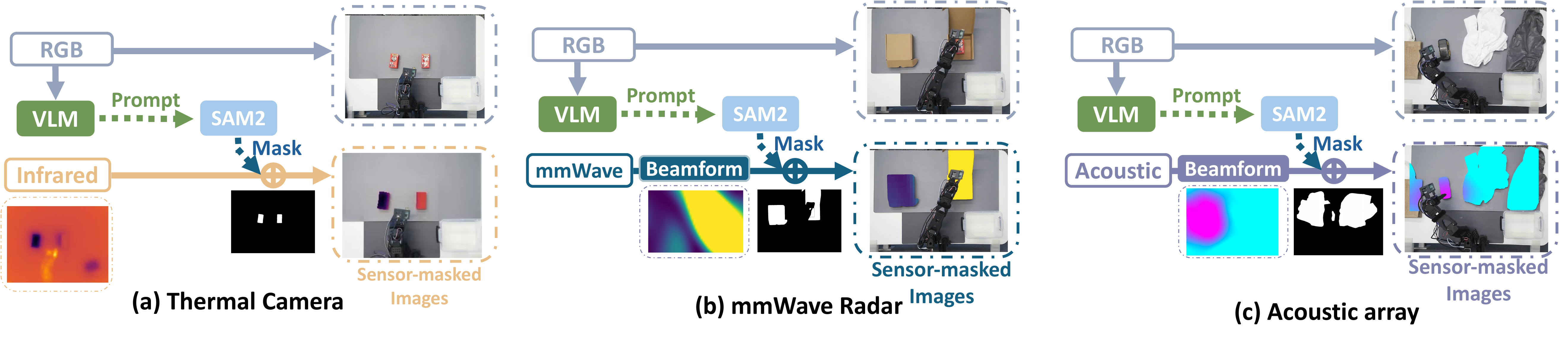}
    \caption{Sensor Data Processing Illustration. We propose a general sensor data processing pipeline applicable to various sensors, including (a) thermal camera, (b) mmWave radar, and (c) acoustic microphone array, by overlaying sensor information on top of RGB images as VLA model input. We update \majorEdit{the} prompt input to the SAM2 model when the task begins and then asynchronously check for updates, so that VLM delay does not affect \majorEdit{the} real-time processing of sensor-masked images.}
    \label{fig:design-sensor-processing}
\end{figure*}

\subsection{Sensor Data Representation}
The first part of our work is sensor-masked image generation, which includes sensor data preprocessing, segmentation mask processing, and sensor-masked image blending. 

\heading{Preprocessing.}
We convert all raw measurements into a \minorEdit{\emph{camera-like 2D spatial representation}}. 
Thermal camera already output\majorEdit{s} raster images—infrared intensity, defined over image coordinates $(u,v)$. 
In contrast, the mmWave radar and the microphone array provide complex signal samples per array element,
$x_{i,k} = A_{i,k} e^{j\psi_{i,k}}$ for the $k$-th element of array $i$, where $i \in \{\text{mmWave}, \text{acoustic}\}$. To obtain inputs with \emph{consistent spatial mapping} (horizontal and vertical viewing axes on a 2D grid), we compute azimuth–elevation heatmaps via conventional delay-and-sum beamforming~\cite{xu2017waveforming}:
\begin{equation}
\begin{aligned}
\mathrm{I}_i(\theta,\phi) &= 20\log_{10}\!\left\lVert \sum_{k=1}^K A_{i,k} e^{j\psi_{i,k}} e^{-j\Phi_{i,k}} \right\rVert, \\
\Phi_{i,k} &= \frac{2\pi}{\lambda_i}\big(x_{i,k}\cos\phi\,\sin\theta + y_{i,k}\sin\phi\big).
\end{aligned}
\label{eq:beamforming}
\end{equation}
where $\theta$ is the azimuth angle, $\phi$ is the elevation angle, $\lambda_i$ is the wavelength received by sensor $i$, $(x_{i,k}, y_{i,k})$ is the position for $k^{th}$ array element (antenna or microphone) in array $i$. One example is shown in Fig. \ref{fig:intro-sensor-motiv}. Similar to the principle of RGB camera, the azimuth-elevation heatmap reveals the environmental information in a direct way for human understanding and acts as the sensor images in the following steps.

\heading{Segmentation and overlay.}
The next step is generating segmentation masks \majorEdit{based on} the task description and RGB images. 
We first send the task request text and RGB image to a vision-language model, GPT-4o~\cite{hurst2024gpt}, to generate the segmentation prompts that describe objects related to the task in the current scene, \majorEdit{such as} `red block/drink', `black phone', \majorEdit{and} `cardboard boxes'. 
\majorEdit{These prompts are initialized when the task begins and are updated periodically to accommodate scene dynamics, such as the introduction of new objects. 
We run these updates in the background at a lower frequency than the robot's primary action-generation loop}, so that VLM output delay does not affect real-time robot manipulation actions.
Then we input the segmentation prompts and RGB image into a segmentation model to generate image masks for task\majorEdit{-}related objects. 
Specifically, we use Grounded SAM 2~\cite{ren2024grounded}, which combines segment anything model~\cite{ravi2024sam} and open-set object detection model, grounding DINO~\cite{liu2024grounding}. 
The final output is a 0-1 matrix, where `1' labels the masked regions for objects. 
The second step can be expressed as:
\begin{equation}
\begin{aligned}
l = \mathrm{VLM}(\mathrm{T}_{\text{task}}, \mathrm{I}_{\mathrm{RGB}}),\\
\text{mask} = \mathrm{SAM2}(l, \mathrm{I}_{\mathrm{RGB}}),
\end{aligned}
\label{eq:segmentation}
\end{equation}
where $l$ represents the segmentation prompt\majorEdit{s} generated by VLM, $\mathrm{T}_{\text{task}}$ is the input task description, and $\mathrm{I}_{\mathrm{RGB}}$ is the RGB image.
\majorEdit{To generate the final sensor-masked images, we first spatially align each sensor modality with the RGB frame through a one-time calibration involving rotation and cropping.
While high-precision pixel matching is not strictly required for the model to learn effectively, this ensures the sensor heatmaps are roughly centered on their physical counterparts.
We then isolate the regions of interest by applying the masks generated in the previous step. This fusion process is formalized as:}
\begin{equation}
\begin{aligned}
\mathrm{I}_{i}^{\mathrm{c}} &= \mathrm{Calibration}(\mathrm{I}_i),\\
\mathrm{I}_{i}^{\mathrm{m}} &= \text{mask}\odot\big(\alpha\,\mathrm{I}_{i}^{\mathrm{c}} + (1-\alpha)\mathrm{I}_{\mathrm{RGB}}\big) \\
&\quad {}+ (1-\text{mask})\odot \mathrm{I}_{\mathrm{RGB}}
\end{aligned}
\label{eq:blending}
\end{equation}
where $i \in \{\text{mmWave}, \text{acoustic}, \text{thermal} \}$ and $\alpha \majorEdit{\in [0, 1]}$ is the hyper-parameter for image blending. The higher $\alpha$ is, \majorEdit{the} more sensor information remains for comprehensive understanding, but less correlation between \majorEdit{the} masked part and \majorEdit{the} unmasked RGB image remains. 
We set $\alpha$ as 1 by default based on our empirical testing.
\majorEdit{These consolidated sensor images then serve as the unified input for our multi-sensor VLA backbone.}

\subsection{Model architecture and training}
The second part is the architecture design and training strategy of our multi-sensory vision-language-action model.

\majorEdit{
Existing VLA models typically encode RGB inputs with a vision encoder, followed by a multi-layer perceptron (MLP) to generate image tokens. 
Simultaneously, the task description is processed by a language tokenizer. 
These tokens are concatenated and fed into a large language model (LLM) backbone, which generates action tokens for robot manipulation.
We reuse and adapt this architecture design. 
}

We present a generalizable and efficient multi-sensory VLA model architecture design as shown in Fig. \ref{fig:design-system-overview}.
We enable effective sensor data understanding by feeding sensor-masked images to \majorEdit{the} existing vision encoder. 
Then for each sensor modality, we input the encoded results into individual multi-layer perceptron (MLP) modules for projection to align with the language tokens. 
The projected tokens for each sensor are concatenated together with language tokens from the task description after passing through \majorEdit{a language tokenizer}. 
\majorEdit{The LLM backbone takes these concatenated tokens as input. 
To generate the final robotic actions, we follow the architecture used in SmolVLA~\cite{shukor2025smolvla} and $\pi_0$~\cite{black2024pi0}, where the LLM backbone is combined with a diffusion or flow matching based Action Expert for robotic action generation. 
We utilize conditional flow matching or diffusion within the action expert to model the continuous distribution of actions, which is inherited from the chosen base model. 
This allows the model to generate a full action chunk step by step from random noise.}
\majorEdit{We use SmolVLA as our base model by default, modify its architecture, and train our model from its pretrained weights. 
Our approach is potentially compatible with various VLA backbones because our sensor-masked image representation allows processing sensor data as standard visual tokens.
}

The overall data flow can be described with:
\begin{equation}
\begin{aligned}
\text{t}_{i} &= \text{MLP}_i\big(\mathrm{E}_I (\mathrm{I}_{i}^{\mathrm{m}} )\big),\\
\text{t}_{\text{task}} &= \mathrm{E}_L(\mathrm{T}_{\text{task}}),\\
\text{action} &= \text{VLA}\big([\text{t}_{1},\,\text{t}_{2},\,\ldots,\,\text{t}_{m},\,\text{t}_{\text{task}}]\big)
\end{aligned}
\label{eq:model}
\end{equation}
where $m$ is the number of sensors we use, $\mathrm{E}_I$ is the image encoder, and $\mathrm{E}_L$ is the language embedding layer, $\text{t}_{i}$ is the embedding of $i$th sensor image. We underline that our architecture does not require all sensors shown here. Instead, we allow flexible sensor setup according to the deployment scenario, budget, etc, for example using a single sensor like \majorEdit{an} infrared camera.  
This design is a general and flexible framework to process beyond-RGB sensor-masked images, compatible with existing Vision-Language-Action models. It means we can utilize the vision-language capability in existing model's pretraining and understand unique features from each sensor easily with a few demonstration data. 

During the model training, \majorEdit{we start from the pretrained base model weight.} We freeze the vision encoder, typical in various Vision-Language-Action models~\cite{shukor2025smolvla, black2024pi0, karamcheti2024prismatic} and Vision-Language models finetuning~\cite{liu2023visual, chen2023minigpt}. Meanwhile, we set all other weights trainable for multi-sensor perception in robotic manipulation. 
\majorEdit{
We initialize the individual MLP projectors for each sensor using weights from the pre-trained RGB projection layer of the base VLA model. 
This strategy leverages established visual feature mappings as a strong prior, allowing the model to adapt quickly to the unique sensor-masked images. 
Finally, we co-fine-tune these sensor-specific MLP modules alongside the unfrozen backbone weights using the collected demonstration datasets.
}

\section{Evaluation}
\label{sec:eval}

We evaluate \name{} with a real-world prototype across several sensor-related manipulation tasks. 
First, we show our unique beyond-RGB perception capabilities, using daily tasks that require non-visible cues, including thermal-based pick-and-place, see-through and open the box containing the object, and uncovering hidden ringing items. 
\name{} significantly outperforms RGB-only VLA models and VLA models trained with unprocessed sensor images.
We also highlight the data efficiency of our approach.
Second, we show that our approach provides superior generalization capability for sensor-related tasks, outperforming baselines.





\begin{figure}
    \centering
    \setlength{\belowcaptionskip}{-10pt}
    \includegraphics[width=0.98\columnwidth]{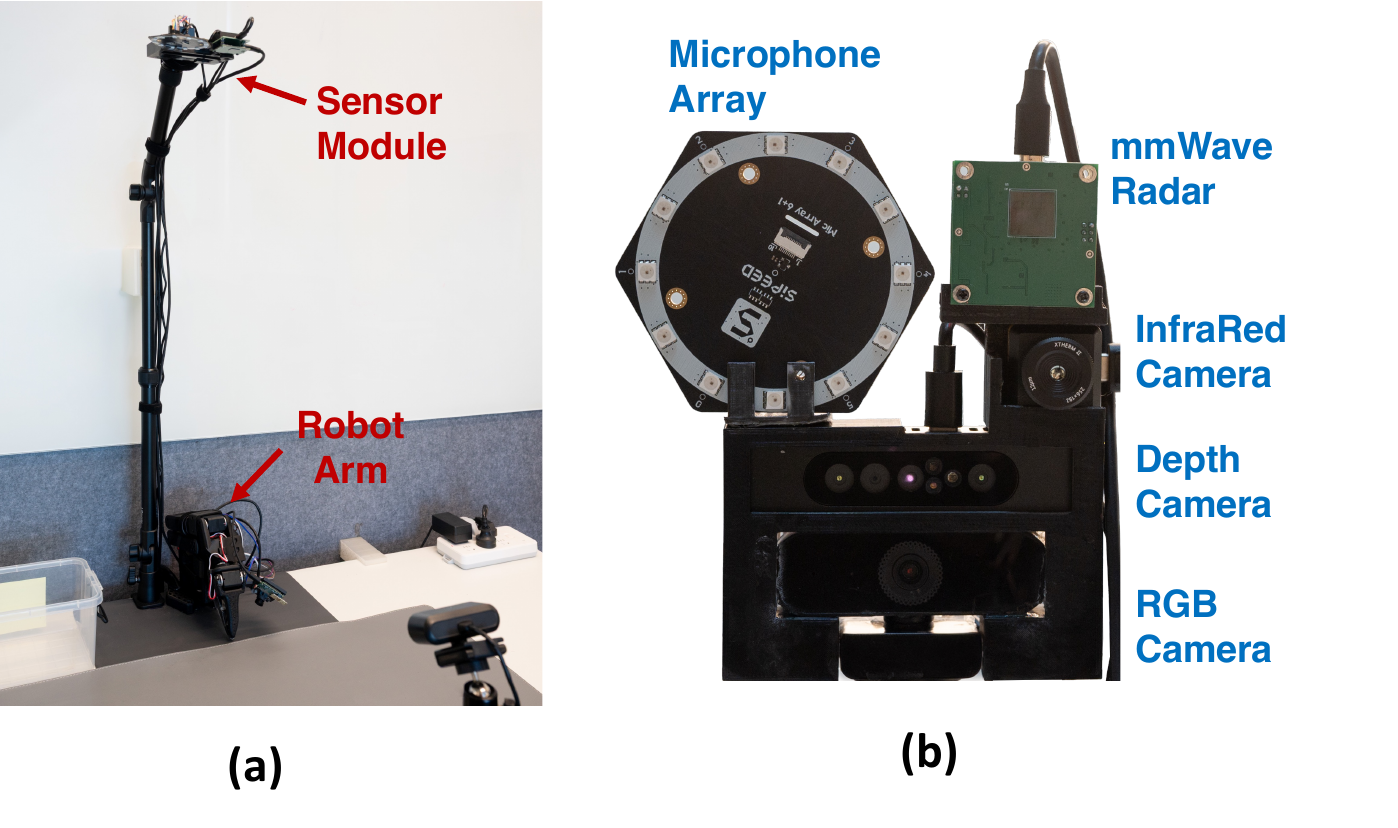}
    \caption{Hardware Implementation. (a) robot arm and sensor setup (b) sensor module, integrating multiple sensors and cameras.}
    \label{fig:eval-hardware-setup}
\end{figure}

\begin{figure}
    \centering
    \setlength{\belowcaptionskip}{-10pt}
    \includegraphics[width=0.98\columnwidth]{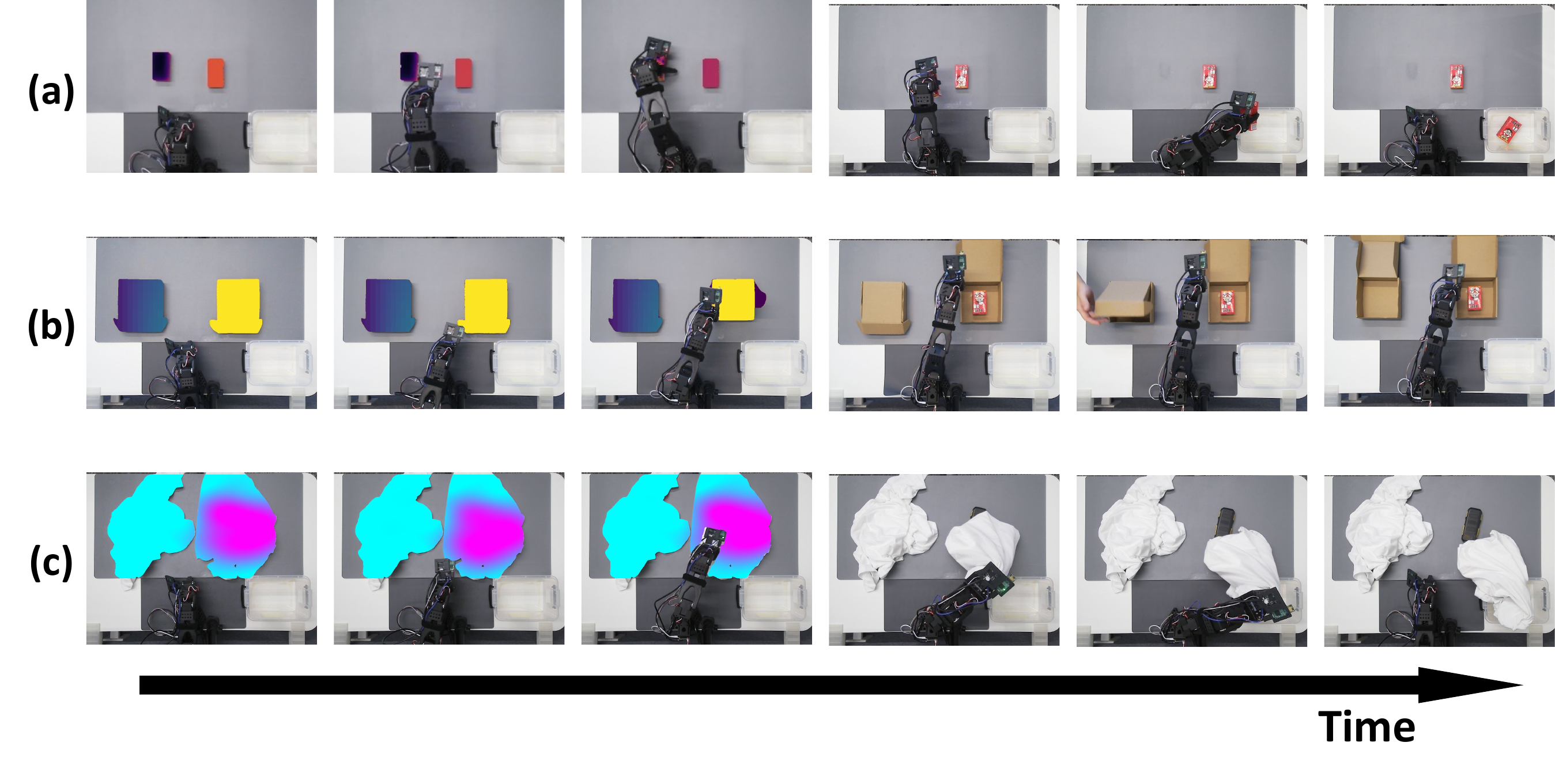}
    \caption{Examples of Robotic Manipulation Task Completion over Time. (a) Thermal: finding the cold drink. (b) mmWave: opening the box with an object inside. (c) Acoustic: uncovering the ringing phone. The first three images are sensor-masked images; the rest of the images are raw RGB images for visibility.}
    \label{fig:eval-task-examples}
\end{figure}

\begin{table*}[t]
\small
\centering
\vspace{2pt}
\setlength{\tabcolsep}{5pt}
\begin{tabular}{lcccc S[table-format=1.2] S[table-format=1.2] S[table-format=1.2] S[table-format=1.2]}
\toprule
 & \multicolumn{4}{c}{Success Rate} & \multicolumn{4}{c}{Task Score} \\
\cmidrule(lr){2-5}\cmidrule(lr){6-9}
 & Thermal & mmWave & Acoustic & \textit{Average} & {Thermal} & {mmWave} & {Acoustic} & {\textit{Average}} \\
\midrule
VLA-RGB  & 28\% & 8\%  & 40\% & 25\% & 0.62 & 0.34 & 0.70 & 0.55 \\
VLA-RAW  & 52\% & 68\% & 48\% & 56\% & 0.74 & 0.84 & 0.62 & 0.73 \\
\myrowcolour
\textbf{\name{}} & \textbf{80\%} & \textbf{84\%} & \textbf{88\%} & \textbf{84\%} & \textbf{0.91} & \textbf{0.88} & \textbf{0.92} & \textbf{0.90} \\
\bottomrule
\end{tabular}
\caption{\name{} success rates and task scores across three types of sensor-modality tasks.}
\vspace{-0.5em}
\label{tab:main-result}
\end{table*}


\subsection{Experimental Setup}
\heading{Implementation.} 
Our prototype includes a SO101 manipulator arm~\cite{cadene2024lerobot} with a standard top-down RGB camera, a front camera, and an arm camera, augmented by our multi-modal sensor suite comprising: a depth camera, an infrared thermal camera, an mmWave radar sensor, and a six-microphone circular array (\figref{fig:eval-hardware-setup}). 
While the RGB camera provides standard visual perception, our additional sensor modalities capture unique physical information that extends perception capability significantly. 

\heading{Model training and inference.}
We use SmolVLA~\cite{shukor2025smolvla} as the base model by default to implement our design and use the pre-trained weights. 
We expect our approach to be compatible with most existing RGB-only VLA models. 
We use \majorEdit{8} NVIDIA A100 GPUs on a server for distributed training and use a local RTX 4090 GPU for model inference during system evaluation. 
\majorEdit{Depending on the size of the demonstration dataset, the time required for training is around 14 hours for 50~K optimization steps with batch size 32.}
For real-time inference, we load both the VLA model and segmentation model on the local RTX 4090 machine, which is able to output 15 predictions per second for end-to-end action prediction. 
\majorEdit{Note that prompting the VLM model for segmentation keywords is invoked at the beginning of the task and only updated infrequently to minimize overhead.}
We expect to have lower delays with code optimization. 

\heading{Task setting.}
We evaluate \name\ on three types of manipulation tasks \majorEdit{that need to leverage non-visual sensory modalities}: 
(1) \textit{Thermal modality}: Distinguishing between a cold and warm drink, picking up the cold drink and placing it into a plastic container;
(2) \textit{mmWave modality}: Seeing through enclosed cardboard/foam (non-metal) boxes with mmWave radar, opening one of the boxes with an object inside and exposing the target object.
(3) \textit{Acoustic modality}: Locating a ringing mobile phone concealed beneath opaque coverings using spatial audio cues from the microphone array, and removing the covering to uncover the phone.
We show examples of successful action trajectories from \name{} in \figref{fig:eval-task-examples}.



\begin{table*}[t]
\small
\centering
\setlength{\tabcolsep}{5pt}
\begin{tabular}{lcccc
  S[table-format=1.2] S[table-format=1.2] S[table-format=1.2] S[table-format=1.2]}
\toprule
 & \multicolumn{4}{c}{Success Rate} & \multicolumn{4}{c}{Task Score} \\
\cmidrule(lr){2-5}\cmidrule(lr){6-9}
 \majorEdit{Base Model} & Thermal & mmWave & Acoustic & \textit{Average}
 & {Thermal} & {mmWave} & {Acoustic} & {\textit{Average}} \\
\midrule
SmolVLA & 80\% & 84\% & 88\% & 84\% & 0.91 & 0.88 & 0.92 & 0.90 \\
Pi0     & 68\% & 60\% & 64\% & 64\% & 0.84 & 0.72 & 0.82 & 0.80 \\
\bottomrule
\end{tabular}
\caption{\majorEdit{\name{} performance when using two different base models as backbone,} evaluated across three types of sensor tasks.}
\vspace{-0.5em}
\label{tab:model-comparision}
\end{table*}



\begin{table*}[t]
\small
\centering
\setlength{\tabcolsep}{6pt}
\begin{tabular}{l cc cc cc}
\toprule
 & \multicolumn{2}{c}{Thermal} & \multicolumn{2}{c}{mmWave} & \multicolumn{2}{c}{Acoustic} \\
\cmidrule(lr){2-3}\cmidrule(lr){4-5}\cmidrule(lr){6-7}
 & Stage 1 & Stage 2 & Stage 1 & Stage 2 & Stage 1 & Stage 2 \\
\midrule
\majorEdit{\name{}-Base}              & \textbf{100\%} & 24\% & 56\% & 40\% & 76\% & 16\% \\
Pretrained VLA-RAW    & 76\% & 84\% & 52\% & 76\% & 60\% & \textbf{92\%} \\
\textbf{Pretrained \name{}} & \textbf{100\%} & \textbf{92\%} & \textbf{92\%} & \textbf{80\%} & \textbf{92\%} & \textbf{92\%} \\
\midrule
\myrowcolour
{Gains} & +0\%/{+24\%} & {+68\%}/{+8\%} & {+36\%}/{+40\%} & {+40\%}/{+4\%} & {+16\%}/{+32\%} & {+76\%}/+0\% \\
\bottomrule
\end{tabular}
\caption{Pretraining effectiveness by comparing per-stage success rates (\%). Gains are improvements over baselines. }
\vspace{-1em}
\label{tab:per_step_SR}
\end{table*}

\heading{Evaluation metrics and baselines.}
We evaluate model performance using task success rates \majorEdit{computed over 25 independent trials per task with random object placement}, complemented by task scores: 0.5 score for choosing the right item to interact with, 0.5 score for performing the correct manipulation, \eg, picking up and placing in the container, opening up the box, \majorEdit{and} removing the coverings. 
For baselines, we compare our approach against \majorEdit{the following ablation baselines:} (1) \textbf{VLA-RGB} \majorEdit{(modality ablation)}: VLA models with standard RGB input only for training and inference, without our architecture changes. \majorEdit{Since we use SmolVLA as the backbone, this baseline model is essentially a SmolVLA model fine-tuned on our task dataset.} (2) \textbf{VLA-RAW} \majorEdit{(representation ablation)}: VLA models with raw sensor data/images for training and inference input. It uses the same model architecture as \name{}, but skips the segmentation and overlay step for the sensor data processing. Note that we still apply beamforming for the VLA-RAW model mmWave and acoustic sensor input to have a strong baseline.

\subsection{Multi-sensory Task Performance}
We first evaluate \name{} performance on manipulation tasks compared with the baselines. Then we evaluate the performance over different foundational robotic models. Finally, we explore the impact of finetuning data amount on final performance. 

\heading{Overall Performance.} 
We evaluate \name{} after training on 100 expert demonstration episodes of thermal and acoustic modality tasks and 200 episodes of mmWave modality task individually. 
The mmWave task requires more training data because opening a box is more difficult to learn compared to typical pick-and-place type of actions, which seldom appears in VLA model pretraining datasets.
For each demonstration, we randomize objects' positions on a table within the robot's workspace.
As the SmolVLA pretraining dataset does not include any non-RGB sensor, we consider the number of episodes reasonable and showing high learning efficiency of our approach.
\tabref{tab:main-result} shows success rates and task scores across tasks.
\name{} finishes the tasks requiring extra sensors successfully at a high rate of 84\% on average.
\name\ consistently outperforms all baseline configurations across the three tasks, demonstrating the effectiveness of our unified multi-sensory perception approach. 
On average, \name{} outperforms VLA-RGB model and VLA-RAW model by {59\%} and {28\% in success rate} respectively. {\name{} also improves the task score by 0.45 and 0.17, respectively.}
This shows that sensor-modalities effectively provide new capabilities for the VLA model and our sensor-masked image representation significantly boosts performance.



 

\heading{Comparing base models.}
\majorEdit{To evaluate the applicability of our approach on different base VLA models, we apply it} to Pi0~\cite{black2024pi0} and compare the performance across 3 tasks. \tabref{tab:model-comparision} shows that both models work, while SmolVLA provides better performance. This is likely due to SmolVLA is pre-trained with lerobot robot arm dataset\majorEdit{~\cite{shukor2025smolvla}}.
Overall, this shows \name{}'s potential for combining with various RGB-only VLA models to provide extra new sensor-modality capabilities without starting from scratch.

\begin{figure}
    \centering
    \setlength{\belowcaptionskip}{-10pt}
    \includegraphics[width=0.95\columnwidth]{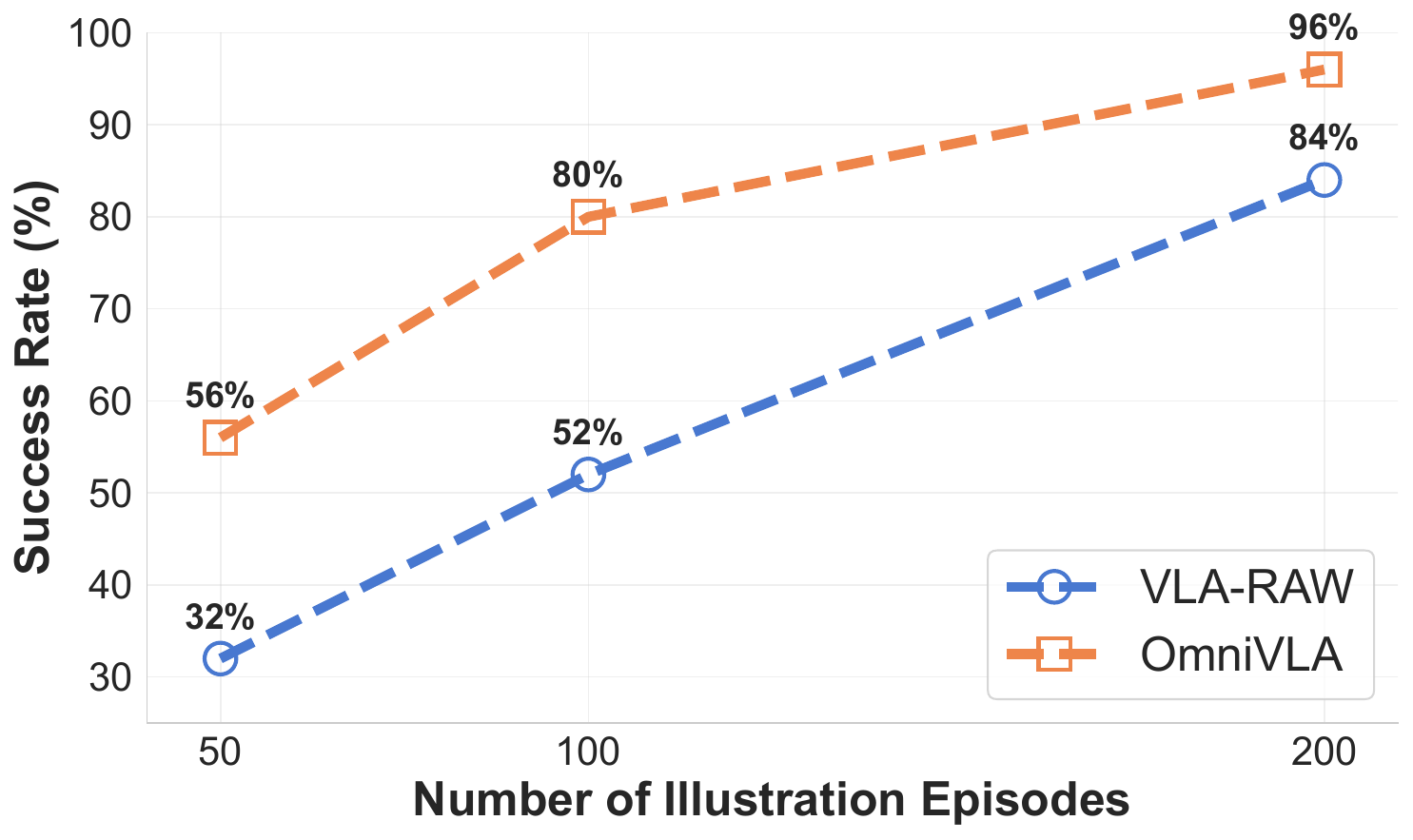}
    \caption{Success rates over number of demonstration episodes.}
    \label{fig:eval-SR_eps_num}
\end{figure}

\heading{Learning efficiency.}
We quantitatively compare the learning efficiency of using sensor-masked images and raw sensor data (no overlaying on RGB images), by training with an increasing number of thermal task episodes.
As shown in \figref{fig:eval-SR_eps_num}, \name{} constantly outperforms VLA-RAW model, achieving similar success rate with only around 50\% of the training episodes. {This shows the data efficiency of our proposed sensor-masked image representation, which significantly reduces the required finetuning data amount compared with using raw sensor images.}

\subsection{Generalization Performance}
Lastly, we evaluate how well our architecture can generalize to unseen tasks by pre-training on a mixed multi-sensory training dataset.

\begin{figure}
    \centering
    \setlength{\belowcaptionskip}{-10pt}
    \includegraphics[width=0.95\columnwidth]{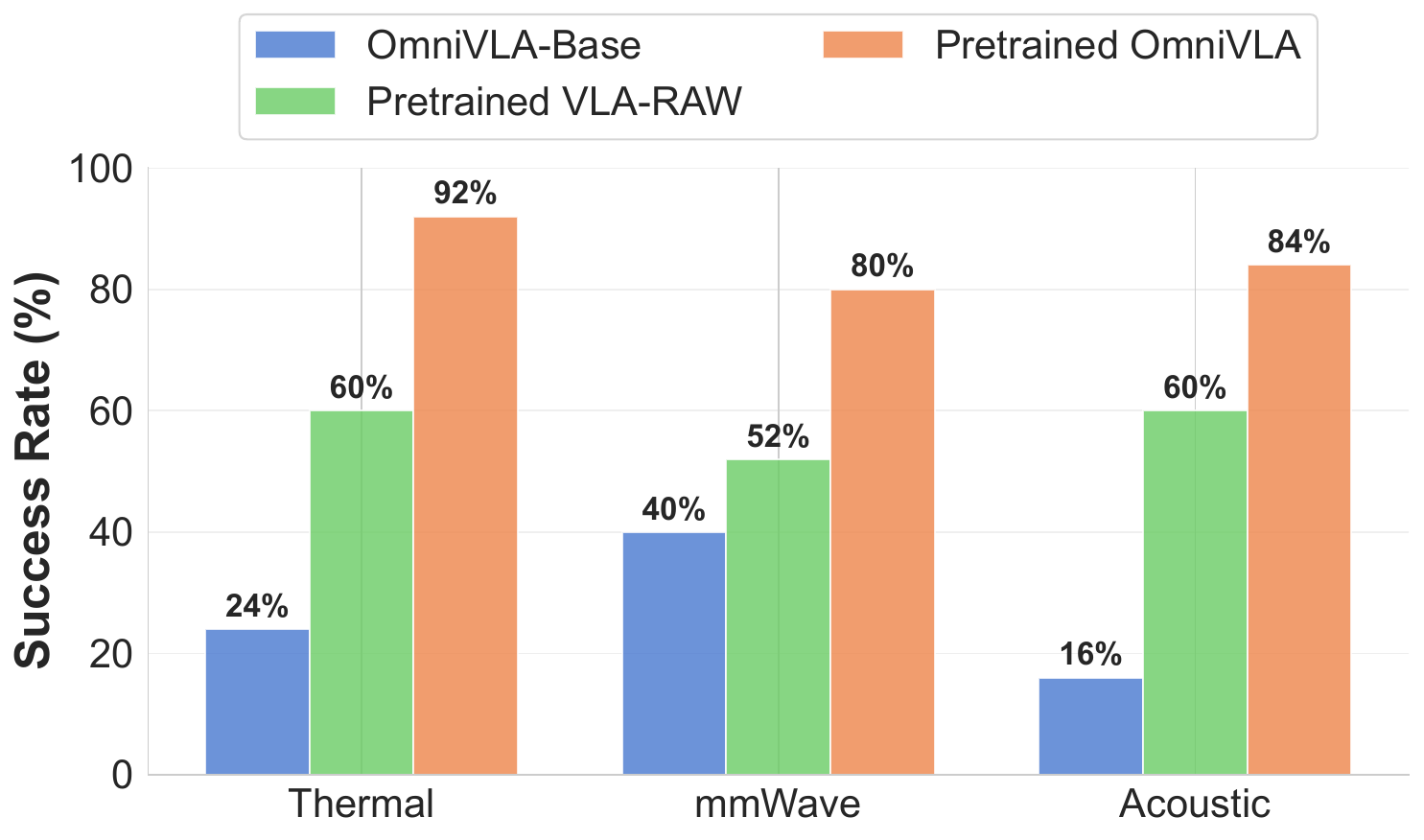}
    \caption{Success rates when adapting to unseen tasks. We compare the pretrained \name{} model with two baselines, \name{}-Base (no pretraining), Pretrained VLA-RAW (pretrained with raw sensor images).}
    \label{fig:eval-SR_gen_type}
\end{figure}

\heading{Multi-sensory pretraining.}
We construct a pretraining corpus of 800 demonstration episodes: 200 episodes for each of three sensor-modality tasks (total 600) and 200 episodes for generic pick-and-place with everyday objects. 

We pre-train our \name{} model with the mixed dataset, then test the performance after performing few-shot learning with only 25 demonstration episodes of performing unseen tasks (matching common RGB-only VLA testing practice).
We induce distribution shifts in object identity and materials based on tasks shown in \figref{fig:eval-task-examples}): (1) swap the drink type, (2) replace cardboard with foam boxes, and (3) substitute T-shirts with towels. The object locations are randomized within the feasible workspace of the robot arm and towels are randomly folded.


\heading{Baselines.} To show the effectiveness, we implement two \majorEdit{ablation} baselines:
(1) \textbf{\name{}-Base} — no multi-sensory pretraining, using \name{} model architecture with SmolVLA backbone weights, finetuned on the 25 demonstrations. 
(2) \textbf{Pretrained VLA-RAW} — \name{} model pretrained on the same 800 episodes but feed raw sensor heatmaps/images directly (no segmentation/overlay), then perform the same 25-shot adaptation.
Our method, \name{}, uses the identical pretraining and few-shot protocols but with sensor-masked images as input.
To compare different methods in more detail, we decompose each task into two stages: \textbf{Stage\,1} (select the correct target to interact with) and \textbf{Stage\,2} (complete the subsequent manipulation). We report stage-wise success rates and scores (\tabref{tab:per_step_SR}), and overall task success rate (\figref{fig:eval-SR_gen_type}).


As shown in \todoEdit{\tabref{tab:per_step_SR}},
\name{} shows substantially better generalization to unseen sensor-modality tasks than both baselines. In {Stage\,1}, \name{} improves success rates by {17\%} over \name{}-Base and by {32\%} over Pretrained VLA-RAW on average. In {Stage\,2}, \name{} yields gains of up to {76\%}. These results suggest two complementary effects: (i) the sensor-masked overlay makes sensor cues spatially aligned with RGB, which simplifies learning the selection policy (Stage\,1); and (ii) large-scale multi-sensory pretraining supplies transferable manipulation priors that boost few-shot control (Stage\,2). 
Consequently, the few-shot task success rate increases by 59\% over \name{}-Base and 28\% over Pretrained VLA-RAW on average, and up to 68\% across the three unseen tasks, as shown in \figref{fig:eval-SR_gen_type}.
To conclude, both sensor-masked images and pretraining significantly improve the success rate on unseen tasks with a few episodes of finetuning, showing powerful generalization capability. 



\section{Conclusion}
\label{sec:conclusion}
We introduce \name{}, a multi-modal vision-language-action model that equips robots with perception capabilities beyond the visible spectrum by integrating sensors, \ie, infrared, mmWave, and acoustic sensors.
Our core design is the {sensor-masked image}, a unified representation that spatially grounds and semantically aligns diverse sensor data onto RGB images. 
This image-native approach allows reusing a pretrained vision encoder, enabling data-efficient learning with lightweight per-sensor projection layers.
Through extensive real-world experiments, we demonstrated that \name{} significantly outperforms baseline models on challenging manipulation tasks, showing an average success rate of \textbf{84\%}, exceeding the 25\% of an RGB-only VLA and the 56\% of a model trained on unprocessed sensor data.
\name{} is a step towards creating more versatile and perceptive robots that can fully understand and interact with their physical surroundings.

\bibliographystyle{IEEEtran}
\bibliography{biblio}

\end{document}